\pdfoutput=1

\documentclass[11pt]{article}

\usepackage{acl}
\usepackage{booktabs}
\usepackage{multicol}
\usepackage{multirow}
\usepackage{times}
\usepackage{latexsym}
\usepackage{amsmath}
\usepackage{amssymb}
\usepackage{xcolor}
\usepackage{pifont}
\usepackage{graphicx}
\usepackage{longtable}
\usepackage[T1]{fontenc}

\usepackage[utf8]{inputenc}

\usepackage{microtype}

\usepackage{inconsolata}

\title{Is Bigger and Deeper Always Better? Probing LLaMA Across Scales and Layers}

\author{
 Nuo Chen$^\clubsuit$
\quad
Ning Wu$^{\diamondsuit}$  
\quad
Shining Liang$^{\diamondsuit}$  
\quad
{\bf Ming Gong$^{\diamondsuit}$}  
 \\
\quad
{\bf Linjun Shou$^{\diamondsuit}$}  
{\bf \quad Dongmei Zhang$^{\diamondsuit}$}  
{\bf \quad Jia Li$^\clubsuit$}\\
\\
  $^\clubsuit$Hong Kong University of Science and Technology (Guangzhou)\\ Hong Kong University of Science and Technology\\
  $^{\diamondsuit}$Microsoft\\
    \texttt{nchen022@connect.ust.hk}, \texttt{jialee@ust.hk}\\}

\begin{document}
\maketitle
\begin{abstract}
This paper presents an in-depth analysis of Large Language Models (LLMs), focusing on LLaMA, a prominent open-source foundational model in natural language processing. 
Instead of assessing LLaMA through its generative output, we design multiple-choice tasks to probe its intrinsic understanding in high-order tasks such as reasoning and calculation. We examine the model horizontally, comparing different sizes, and vertically, assessing different layers.
We unveil several key and uncommon findings based on the designed probing tasks: (1) Horizontally, enlarging model sizes almost could not automatically impart additional knowledge or computational prowess. Instead, it can enhance reasoning abilities, especially in math problem solving, and helps reduce hallucinations, but only beyond certain size thresholds; (2) In vertical analysis, the lower layers of LLaMA lack substantial arithmetic and factual knowledge, showcasing logical thinking, multilingual and recognitive abilities, with top layers housing most computational power and real-world knowledge.  These findings provide new observations into LLaMA's capabilities, offering insights into the current state of LLMs. To reproduce our results and access datasets, please refer to \url{https://github.com/nuochenpku/LLaMA_Analysis}.
\end{abstract}

\section{Introduction}

Large language models (LLMs) \cite{openai2023gpt4, scao2022bloom, chen-2023-large,  yao2022react, chen2023breaking} have shown significant potential in numerous high-order open-generation tasks such as mathematical and logical reasoning.
 LLaMA \cite{llama2}, an open-source, state-of-the-art foundational large language model has been designed to facilitate research in natural language processing communities. In a relatively brief period, LLaMA has garnered significant attention. This prominence can be attributed to its inherent accessibility and demonstrated efficacy across a diverse array of text-generation tasks \cite{hu2021lora, chen2022would, chen2023orca, gao2023pal}.
\begin{figure}[!t]
\centering
\includegraphics[width=0.95\linewidth]{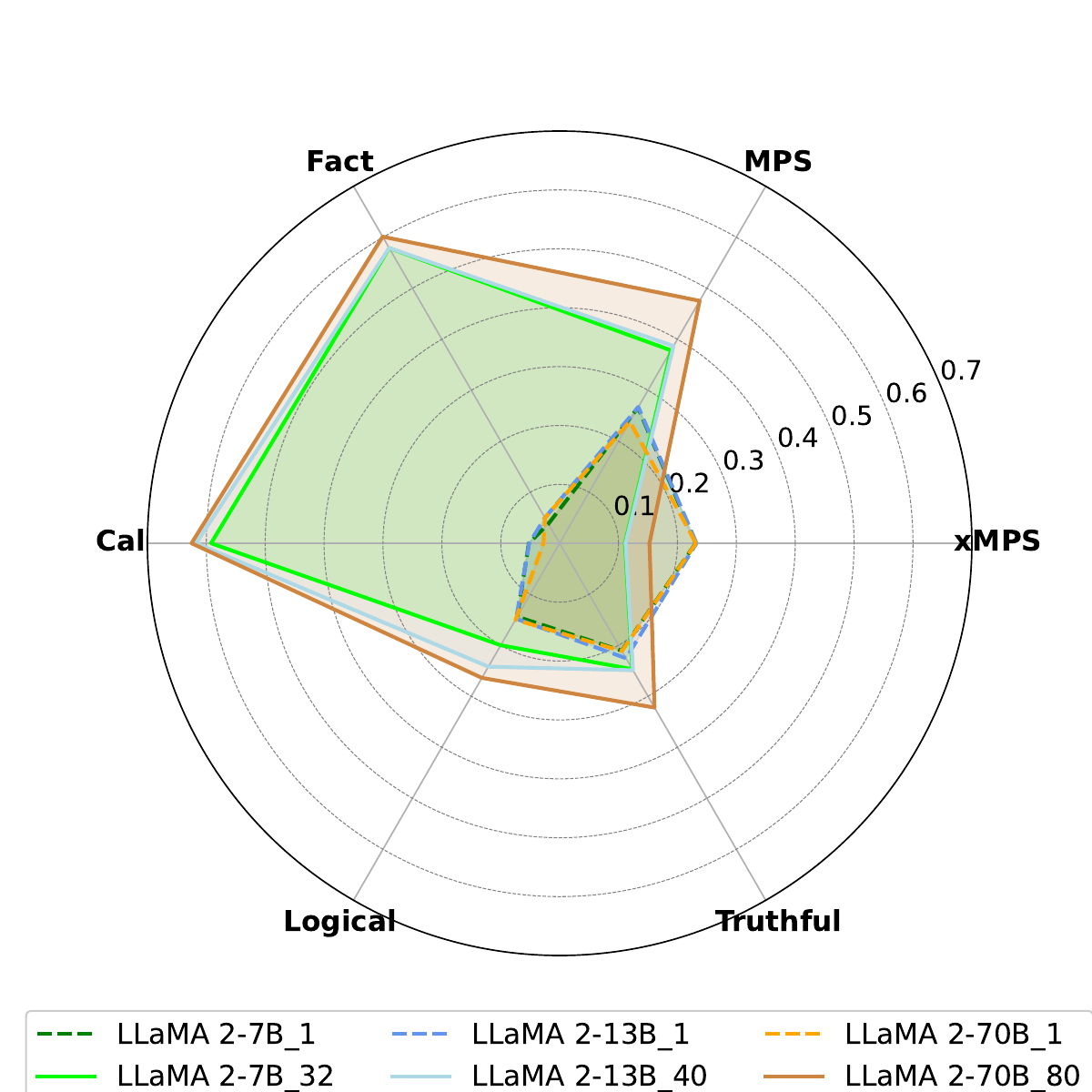}
\caption{Overall Comparison with LLaMA 2 7B-70B in our probing tasks. Detailed introduction of each task include in Section \ref{sec: probing}. Dashed lines represent the first layer of each model, while solid lines represent the last layer of the model.
}
\label{fig:framework}
\vspace{-10pt}
\end{figure}
Beyond LLaMA's impressive generative capabilities, can we further uncover its intrinsic understanding abilities? Does bigger and deeper always lead to better performances in  its advanced capabilities such as computational and reasoning sensitivity? Addressing this question is not only instrumental in comprehending the foundations of its success, but it also facilitates an understanding of its inherent limitations. This, in turn, can guide future advancements in the architecture and training optimization of LLMs.

In this paper, we conduct a series of experiments to probe the nature of LLaMA on five higher-order tasks under in-context learning, including \textit{calculation}, \textit{math problem solving} (MPS), \textit{logical reasoning}, \textit{truthfulness}, and \textit{factual knowledge detection}. The latter two are considered as the important symbols of hallucination.
In these tasks, we probe the model's capabilities from two distinct perspectives: 1) \textbf{Horizontally}: Comparing the model's abilities across different sizes (Scaling Law); 2) \textbf{Vertically}: Comparing the different layers capabilities of the same size model (Layer-wise). Instead of directly testing LLMs via their open-text generation abilities, as is usually done, we prob LLaMA with a set of challenging multiple-choice questions. The primary considerations for this design are: Firstly, it offers controlled and efficient evaluation, with clear, quantifiable results that reduce ambiguity.  This approach allows for directly targeted testing of specific knowledge areas and reasoning skills, as well as validating models' sensitivity to correct or incorrect answers; Secondly, our experimental observations reveal a tendency for LLaMA's lower layers to produce repetitive words rather than coherent sequences, which would lead to an unfair layer-wise comparison.

In the context of our experiments corroborating each other, we draw the following conclusions:

Horizontally: (1) The primary benefit of increasing model size lies in the enhanced reasoning abilities of the models, most notably in their improved capacity in MPS. This increase in size also tends to reduce the occurrence of hallucinations. However, these improvements are only evident when certain LLM size thresholds are surpassed, known as emergent abilities \cite{DBLP:journals/tmlr/WeiTBRZBYBZMCHVLDF22}. For instance, models ranging from 7B to 13B show comparable performance across all probing tasks. It's only when the model size increases from 13B to 70B parameters that a noticeable improvement in reasoning capabilities and a reduction in hallucination issues can be observed, as shown in Figure \ref{fig:framework}; (2) The pure arithmetic capabilities and inherent factual knowledge of LLaMAs with different parameter sizes are remarkably similar. In other words, increasing the model size does not necessarily impart additional factual knowledge or significantly enhance computational capabilities, especially when the same volume of pre-training corpus is used.

Vertically: (1) We find that the lower and middle layers of LLaMA have almost negligible pure arithmetic and factual knowledge capabilities. As the network layers deepen, there is a noticeable leap in performance. Contrarily, even at the very lowest layers, LLaMA possesses logical thinking and recognitive abilities, such as in mathematics, logical reasoning, and avoiding hallucinations. While these abilities do enhance slightly with deeper layers, the improvement remains quite limited. \textit{This implies that current LLMs predominantly house computational power and real-world knowledge in their upper layers, while the lower layers are geared towards relevant abstract thinking but lack substantial real-world knowledge and computational skills.} (2) Interestingly, in our layer-by-layer performance comparisons, we observe that the model's optimal performance in MPS and computational abilities is not always at the final layer. More often, these peak capabilities are found in several layers before the last. However, in contrast, for representing factual knowledge, the final layer of the model proves to be exceptionally crucial.

Further, we extend the mathematical probing tasks to the cross-lingal reasoning context. Specifically, we maintain the questions and incorrect options unchanged and translate the correct answers into other languages to assess the LLaMA's multilingual proficiency. In this setting, our layer-wise experiments show an effect completely opposite to monolingual reasoning: models' performance gradually decreases as the layers deepen. \textit{This indicates that  LLaMA's earlier layers are responsible for preserving general multilingual features.}

Of note, the results presented in our experiments do not necessarily equate directly to the generative capabilities of LLaMA. Rather, in this paper, we provide a novel and comprehensive perspective for observing the natural performance of LLaMA, giving insights to understand the current LLMs better.

\section{LLaMA}
LLaMA \cite{touvron2023llama, touvron2023llama2}  is a series of foundation large language models, released by META, has becomes the most popular open-source LLMs in NLP communities. LLaMA is built on transformer layers \cite{DBLP:conf/nips/VaswaniSPUJGKP17}, trained on trillion of tokens with the language modeling objective, showing powerful abilities down-stream tasks. The contextualized representations are optimized by predicting the next token based on the input sequences.

In this work, we probe   LLaMA 2 series LLMs with our designed tasks, ranging from 7B to 70B parameters in \textbf{in-context learning}. Concretely, LLaMA 2-7B, 13B and 70B consist of 32, 40 and 80 transformer layers with 4096, 5120 and 8192 hidden embedding sizes, separately.

\section{Probing Tasks}
\label{sec: probing}

\begin{table}[t]
    \begin{center}
    \centering
    \small
    \begin{tabular}{l|ccccccc}
    \toprule
   \multirow{1}{*}{\textbf{Type}} & \textbf{Bit}
 & \textbf{+} & \textbf{-} & \textbf{\(\times\)} &\textbf{$\div$} & \textbf{Mix-2} & \textbf{Mix-3} \\
    \midrule

\multirow{3}{*}{\textbf{Int}} & 1-2& 200 &200&200 &200&200 &200\\
 & 3-4& 200 &200&200 &200&200 &200\\
 & 5-6& 200 &200&200 &200&200 &200\\
\midrule
\multirow{3}{*}{\textbf{Float}} &  1-2&200 &200&200 &200&200 &200 \\
 &  3-4&200 &200&200 &200&200 &200 \\
 &  5-6&200 &200&200 &200&200 &200 \\

    \bottomrule
    \end{tabular}
    \end{center}
    \caption{
 Test data statistics in our arithmetic tasks. }
    \vspace{-10pt}
    
    \label{tab:calculation}
\end{table}

\begin{table*}[!t]\footnotesize
\centering
\small
\vspace{-5mm}
\begin{tabular}{l|p{0.85\linewidth}}
\toprule
\textbf{Task Type} & \textbf{Query \(\&\) Options} \\
\midrule
\multirow{2}{*}{\textbf{Arithmetic-Int}} & \textbf{Query}: 2331 + 2693 = ? \quad \textbf{Options}: \textbf{5024} ({\color{green}\ding{51}}); 5018; 5005; 5025 \\

 & \textbf{Query}: 109848 \(\div\) 199 = ? \quad \textbf{Options}: \textbf{552.0} ({\color{green}\ding{51}}); 516.0; 558.0; 567.0 \\
\midrule
\multirow{2}{*}{\textbf{Arithmetic-Flo}} & \textbf{Query}: 7.682 + 28.894 = ?
 \quad \textbf{Options}: \textbf{36.576} ({\color{green}\ding{51}}); 28.576; 40.909; 38.076 \\
& \textbf{Query}: 25.204 \(\times\) 88.29 \(\div\) 12.133 = ?
\quad \textbf{Options}: \textbf{183.406} ({\color{green}\ding{51}}); 183.739; 185.406; 181.962 \\
\midrule
\multirow{6}{*}{\textbf{MPS-Cal}} & \textbf{Query}: Peyton has 3 children and they each get a juice box in their lunch, 5 days a week.  The school year is 25 weeks long.  How many juices boxes will she need for the entire school year for all of her children? \\
\cmidrule{2-2}
& \textbf{Options}: \textbf{Peyton needs 25 weeks x 5 days x 3 children = 375 juice boxes} ({\color{green}\ding{51}}); \\
& 25 weeks x 5 days x 3 children = 75 juice boxes; \\
& Given the conditions of the problem, 3 children, 5 days a week, 25 weeks long, that's 3*5*25 = 105 juice boxes needed. \\

\midrule
\multirow{13}{*}{\textbf{MPS-Rea}} & \textbf{Query}: A family of 12 monkeys collected 10 piles of bananas. 6 piles had 9 hands, with each hand having 14 bananas, while the remaining piles had 12 hands, with each hand having 9 bananas. How many bananas would each monkey get if they divide the bananas equally amongst themselves? \\
\cmidrule{2-2}
& \textbf{Options}: \textbf{The first 6 bunches had 6 x 9 x 14 = 756 bananas. There were 10 - 6 = 4 remaining bunches. The 4 remaining bunches had 4 x 12 x 9 = 432 bananas. All together, there were 756 + 432 = 1188 bananas. Each monkey would get 1188/12 = 99 bananas} ({\color{green}\ding{51}}); \\
& 6 piles had 6 x 9 x 14 = 756 bananas. The remaining 6 piles had 6 x 12 x 9 = 648 bananas. All together, there were 756 + 720 = 1476 bananas. Each monkey would get 1476/12 = 123.0 bananas; \\
& 6 piles had 6 x 9 x 14 = 756 bananas. There were 10 - 6 = 4 piles of bananas with 12 hands and 4 piles of bananas with 6 hands. The 4 piles of bananas with 12 hands had 4 x 12 x 9 = 432 bananas. The 4 piles of bananas with 6 hands had 4 x 6 x 9 = 216 bananas. There were 756 + 432 + 240 = 1428 bananas. Every monkey will get 1428/12 = 119.0 bananas \\
\bottomrule
\end{tabular}
\caption{ Testing examples in our designed calculation and MPS probing tasks. }
\label{table:examples}
\end{table*}

\begin{table*}[t]
    \begin{center}
    \centering
    \small
    \begin{tabular}{l|cccccc}
    \toprule
  \textbf{Tasks} & \textbf{Arithmetic-Int/Float} &  \textbf{Reclor}&\textbf{(x) MPS-Cal} &\textbf{(x) MPS-Rea} & \textbf{TruthfulQA} & \textbf{LAMA}$^*$\\
    \midrule
Avg. Ground-truth & 1 &1 &1& 1& 3.5&1\\
Avg. Candidates & 4 & 4 & 3 & 4.9 &7.6&9.7\\
Total Queries & 3600& 500 & 712 & 1000&817 &3070\\
    \bottomrule
    \end{tabular}
    \end{center}
    \caption{
Overall Test data statistics in our probing tasks. $\textbf{LAMA}^*$ refers to we only use s subset of original corpus. }
    \vspace{-10pt}
    
    \label{tab:mps}
\end{table*}

Probing tasks are generally utilized to explore the inherent knowledge and linguistic features within deep learning models. Previously, \citet{DBLP:conf/acl/JawaharSS19} employed a series of probing tasks to examine the internal representations of BERT. However, with the rapid advancement of LLMs, there currently lacks comprehensive research that deeply analyzes the relationship between the higher-order capabilities of contemporary LLMs and factors such as model size and network layers.

To bridge this gap,   we use probing tasks
to access LLaMA in their ability to encode different types of features across two views: model size and individual layers. Specifically, we devise five high-order tasks: \textit{calculation}, \textit{math problem solving} (MPS), \textit{logical reasoning}, \textit{truthfulness}, and \textit{factual knowledge detection}. We include the latter two as the \textit{hallucination detecting} in the following.
Besides, we also probe LLaMA efficiency in \textit{multilingual mathematical reasoning}. In this section, we will illustrate them sequentially.

\subsection{Calculation}
In this paper, we focus on testing LLMs in basic arithmetic tasks, including four simple arithmetic expressions: addition (+), subtraction (-),  multiplication (×) and division ($\div$):

\begin{itemize}
    \item Add  of two elements within 1$\sim$100, 100$\sim$10000, 10000$\sim$100000, separately.
    \item Subtract  of two elements within 1$\sim$100, 100$\sim$10000, 10000$\sim$1000000, separately.
    \item Multiply  of two elements within 1$\sim$100, 100$\sim$10000, 10000$\sim$1000000, separately.
    \item Division  of two elements within 1$\sim$100, 100$\sim$10000, 10000$\sim$1000000, separately.
    \item Complex arithmetic operations that require performing \textbf{two} operations of addition, subtraction, multiplication, or division.
    \item Complex arithmetic operations that require performing \textbf{three }operations of addition, subtraction, multiplication, or division.

\end{itemize}

Of note, the elements used in the above arithmetic operations include \textbf{integers} and \textbf{floating-point numbers} (with precision up to three decimal places), separately. Table \ref{tab:calculation} shows the corresponding data statistics. Since we probe the computational abilities of LLaMA through the multiple-choice question answering task, to increase the difficulty and test the model's sensitivity to minor differences in computational results, we randomly add or subtract a floating-point number within $\mathbf{\pm20}$ (except 0) to the correct answer to create three different but indistinct incorrect options.

This design of our test set allows for an intuitive and fine-grained comparison of 1) the model's relative strengths and weaknesses in addition, subtraction, multiplication, and division operations; 2)  the model's performance patterns when faced with complex calculations; 3)  the variations in the model's computational abilities when dealing with floating-point numbers and integers, 1-2 digit, 3-4 digit, 5-6 digit numbers respectively. Our data are constructed by calling python \texttt{random.randint()} and \texttt{random.uniform()} functions.

\subsection{Math Problem Solving}
\label{MPS}

Besides validating LLaMA in arithmetic tasks, we also test the model in MPS tasks to comprehensively review its math reasoning abilities.

We select GSM8K \cite{DBLP:journals/corr/abs-2110-14168} as our source data to construct challenging and misleading options that effectively fool the model. Our strategy involves the following steps:

\begin{itemize}
    \item  We first fine-tune the LLaMA 2-13B model on GSM8K, and then perform rejection sampling via inference 100 times to generate various reasoning paths based on the resulting model.
    \item Next, we extract all the formulas in each reasoning path and validate their accuracy. We use the erroneous reasoning paths to construct our probing task data:
    \begin{itemize}
        \item If a reasoning path only contains computational errors, meaning the correct answer can be obtained by recalculating, we retain it as part of our \textbf{MPS-Cal} probing test set.
        \item  If all computations in a reasoning path are correct, but the final conclusion is wrong, indicating a reasoning error, we use it for our \textbf{MPS-Rea} test set. 
    \end{itemize}
\end{itemize}
 The MPS-Cal focuses on assessing the model's sensitivity to computational results  in solving mathematical problems. Conversely, MPS-Rea emphasizes evaluating the model's ability to discern correct from incorrect reasoning paths, requiring a superior level of understanding and reasoning capabilities. Table \ref{table:examples} shows several examples in MPS and calculation tasks.

 \subsection{Logical Reasoning}

As a key indicator of the advanced capabilities of contemporary LLMs, logical reasoning stands out for its importance in examining, analyzing, and critically assessing arguments in natural language. In our study, we employ Reclor \cite{DBLP:conf/iclr/YuJDF20} as a testing platform to evaluate the logical reasoning skills of these large models. Reclor comprises a dataset derived from logical reasoning questions found in standardized tests for graduate admissions. Each sample from Reclor contains one context, one corresponding question
and four options.

\subsection{Hallucination  Detecting}

Hallucination, which means generating content that deviates from real-world facts observed during pretraining, is considered one of the most challenging issues in LLMs. In order to further investigate the relationship between hallucination and model layers and size, we conduct tests from two aspects: 1) Measure whether a language model is truthful in generating answers to questions, also known as truthfulness; 2) Test the model's internal factual knowledge. We use TruthfulQA MC tasks \cite{DBLP:conf/acl/LinHE22} and LAMA \cite{DBLP:conf/emnlp/PetroniRRLBWM19} as test beds for these two aspects, respectively. It is important to note that in TruthfulQA, there may be more than one correct answer, accompanied by 4-5 incorrect options. As for LAMA, we randomly extract a subset containing 3070 questions along with their 9-10 corresponding options. Table \ref{tab:mps} presents detailed data statistics in our probing tasks.

\begin{table*}[t]
    \begin{center}
    \centering
    \small
    \begin{tabular}{ccccccccc}
    \toprule
  \multirow{2}{*}{\textbf{Model Size}} &  \multirow{2}{*}{\textbf{LAMA}$^*$} &  \multirow{2}{*}{\textbf{Reclor}}&\multirow{2}{*}{\textbf{MPS-Cal}} & \multirow{2}{*}{\textbf{MPS-Rea}} & \multicolumn{2}{c}{\textbf{TruthfulQA}} &  \multicolumn{2}{c}{\textbf{Arithmetic}}\\
  \cmidrule{6-7} \cmidrule{8-9}
  &(\texttt{Fact})& (\texttt{Logical})& &&MC1&MC3& Int & Float\\
    \midrule
\textbf{7B} &57.9& 20.0 &28.7 &47.0& 28.6& 20.7 & 67.9 & 52.5\\
\textbf{13B} &57.9& 23.7 & 30.2 & 46.6 & 29.1 &20.7 & 70.6 & 52.6 \\
\textbf{70B} &58.7 & 26.4& 48.3 & 51.9 & 37.3&27.1&70.8 & 52.9\\
    \bottomrule
    \end{tabular}
    \end{center}
    \caption{
Overall performances of each size LLaMA 2 model in our probing tasks. $\textbf{LAMA}^*$ refers to we only use s subset of original corpus. MC3 accuracy means the normalized total probability assigned to all true answers among candidates in TruthfulQA. }
    \vspace{-10pt}
    
    \label{tab:model_size}
\end{table*}

\subsection{Cross-Lingual Math Problem Solving }

In this study, we delve further into LLaMA's multilingual abilities. We translate the correct answers from the collected two datasets: MPS-Cal and MPS-Rea in Section \ref{MPS} into four additional languages: \textit{Chinese}, \textit{French}, \textit{Spanish}, and \textit{Thai}, while keeping the questions and other incorrect options as they are, the new resulting test sets named \textbf{xMPS-Cal} and \textbf{xMPS-Rea}. This setting offers several advantages: Firstly, it tests the model's capacity in cross-lingual reasoning transfer, demonstrating its proficiency in not just recognizing but also reasoning in multiple languages. Secondly, by mixing incorrect choices with correct answers in different languages, we robustly assess the model's adaptability and comprehension across linguistic barriers. This unique setup challenges the model's ability to process and integrate multilingual information, not only evaluates the model's language-specific capabilities but also its overall versatility in cross-lingual understanding and reasoning.

\subsection{Test Setting}
Consider a probing dataset \(D\) = \(\{Q, C, O\}\), where \( Q \), \(C\) and \(O\) denote a set of questions, contexts  (only exits for LAMA), and answer options.  For each question \( q \in Q \), there is a corresponding set of answer choices, denoted as \( o \in O \), where \( o = \{o_1, o_2, ..., o_{n-1}, a\} \), \( n \) is the number of answer choices, and \(a\) refers to the correct answer.

The model's task is to identify the correct answer from the set \( o \) for each question \( q \). 
It need to
assign the highest log-probability of completion following the question, independent of the other answer choices \cite{chuang2023dola}. This selection process can be mathematically represented as:
\begin{align}
    o_i^* &={\mathrm{argmax}} \, \log P(o_i | q) \\
\text{Acc} &= 
\begin{cases} 
1, & \text{if } a^* > (o_1^*,..o_{n-1}^*), \\
0, & \text{otherwise}.
\end{cases}
\end{align}



Where \( \log P(o_i | q) \) is the log-probability that the choice \( o_i \) to question \( q \), as evaluated by the model.


\subsection{Experimental Settings}
We select LLaMA 2 from 7B to 70B as our experimental subject. Observing that LLaMA 2 exhibits significant instability in zero-shot testing, we choose to implement few-shot prompting in our probing tasks to optimize the model's performance. In TruthfulQA and LAMA, we respectively employ 6-shot (Table \ref{table:btfqa_prompt}) and 4-shot (Table \ref{table:fact_prompt}) approaches. For reasoning tasks, we consistently use 4-shot for both (x) MPS (Table \ref{table:mps_prompt}) and logical reasoning (Table \ref{table:logical_prompt}). In calculation tasks, we use 6-shot examples (Table \ref{table:cal_prompt}). Detailed prompts are presented in Appendix \ref{prompts}.

\section{Experiments on Probing Model Size}
In this section, we are dedicated to presenting a comparison of the results from LLaMA of different sizes on our probing tasks, as shown in Table \ref{tab:model_size}\footnote{Of note, we use the model last layer to count its  performances in this section.}.  
In Table \ref{tab:cal_results}, we showcase the detailed performance of models under different arithmetic rules and digit counts. Combining these two tables, we can draw the following conclusions:

 \paragraph{Increasing model size hardly enhance the model's internal knowledge.} From Table \ref{tab:model_size}, we can see that the performance of LLAMA 2-7B and 13B on LAMA is identical , and even increasing the model size to 70B results in only a slight improvement (58.7\% vs. 57.9\%). This indicates that only increasing model size is difficult to improve the model's ability to remember and understand knowledge present in the training corpus, provided the training data remains the same.

\begin{table}[t]
    \begin{center}
    \centering
    \small
    \resizebox{0.95\columnwidth}{!}{
    \begin{tabular}{l|ccccccc}
    \toprule
  \textbf{Size} & \textbf{Bit}
 & \textbf{+} & \textbf{-} & \textbf{\(\times\)} &\textbf{$\div$} & \textbf{M-2} & \textbf{M-3} \\
    \midrule
\multicolumn{8}{c}{Integer Arithmetic} \\
\midrule
 \multirow{3}{*}{\textbf{7B}} &1-2& 99.5 &100.0&95.0 &100.0&55.5 &39.5\\
 & 3-4& 98.0 &98.0&59.0 &59.5&48.5 &20.5\\
  &5-6& 89.0 &83.0&53.0 &30.5&48.5 &17.5\\
 \cmidrule{2-8}
  \multirow{3}{*}{\textbf{13B}} &1-2& 99.5 &100&98.5 &100.0&58 &44.5\\
 & 3-4& 99.5 &99.0&73.5 &69.5&53.5 &26.0\\
  &5-6& 96.5 &96.5&63.5 &25.5&53.5 &17.5\\
  \cmidrule{2-8}
   \multirow{3}{*}{\textbf{70B}} &1-2& 99.5 &100.0&98.0 &100.0&64.0 &46.0\\
 & 3-4& 99.0 &99.0&75.0 &68.0&42.0 &18.0\\
  &5-6& 100.0 &100.0&77.0 &23.0&41.0 &20.0\\
\midrule
\multicolumn{8}{c}{Floating-point Arithmetic} \\
\midrule
\multirow{3}{*}{\textbf{7B}} &1-2& 99.5 &100&23.0 &78.0&37.0 &30.0\\
 & 3-4& 98.0 &98.0&18.0 &17.0&32.0 &19.0\\
  &5-6& 94.0 &87.0&19.5 &13.0&35.0 &14.5\\
 \cmidrule{2-8}
  \multirow{3}{*}{\textbf{13B}} &1-2& 99.0 &100.0&26.0 &90.0&40.0&28.0\\
 & 3-4& 99.5 &99.5&14.5 &20.5&33.0 &19.0\\
  &5-6& 99.0 &96.5&13.5 &13.5&39.5 &17.0\\
  \cmidrule{2-8}
   \multirow{3}{*}{\textbf{70B}} &1-2& 100.0 &100.0&26.5 &99.0&43.0 &30.0\\
 & 3-4& 98.5 &100.0&14.0 &39.0&39.5&17.0\\
 &5-6& 98.5 &99.5&14.5 &17.5&45.5 &17.5\\

    \bottomrule
    \end{tabular}
    }
    \end{center}
    \caption{
 Detailed results of different operations in our probing arithmetic tasks. \textbf{M-2/3} refers to arithmetic expression that requires 2/3 times mix operations.}
    \vspace{-10pt}
    
    \label{tab:cal_results}
\end{table}

\begin{figure}[!t]
\centering
\includegraphics[width=0.95\linewidth]{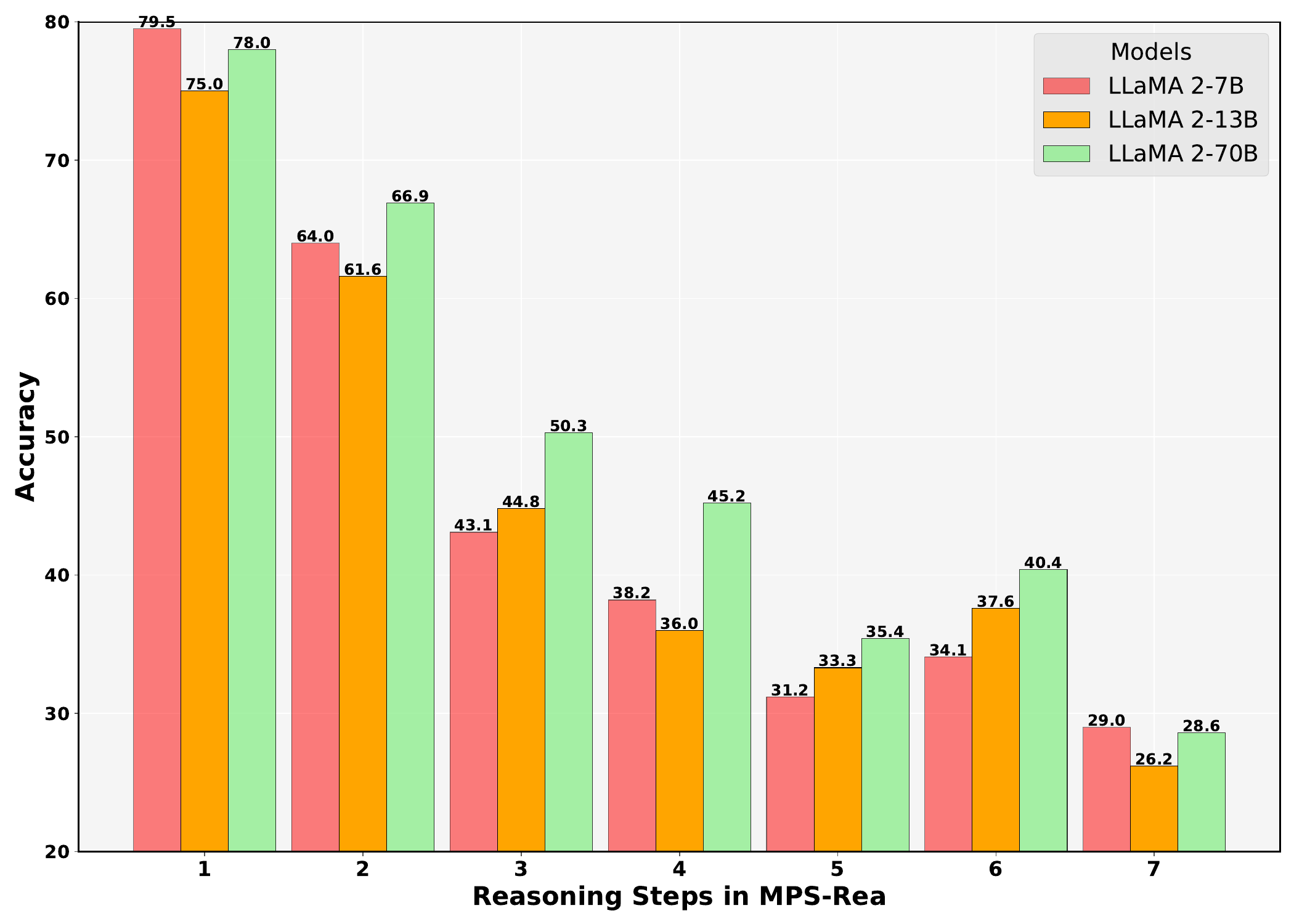}
\caption{Overall comparison between LLaMA 2 7B to 70B dealing with different reasoning steps problems in our probing MPS-Rea tasks. 
}
\label{fig:model_rea}
\vspace{-10pt}
\end{figure}

\begin{figure*}[!t]
\centering
\includegraphics[width=0.95\linewidth]{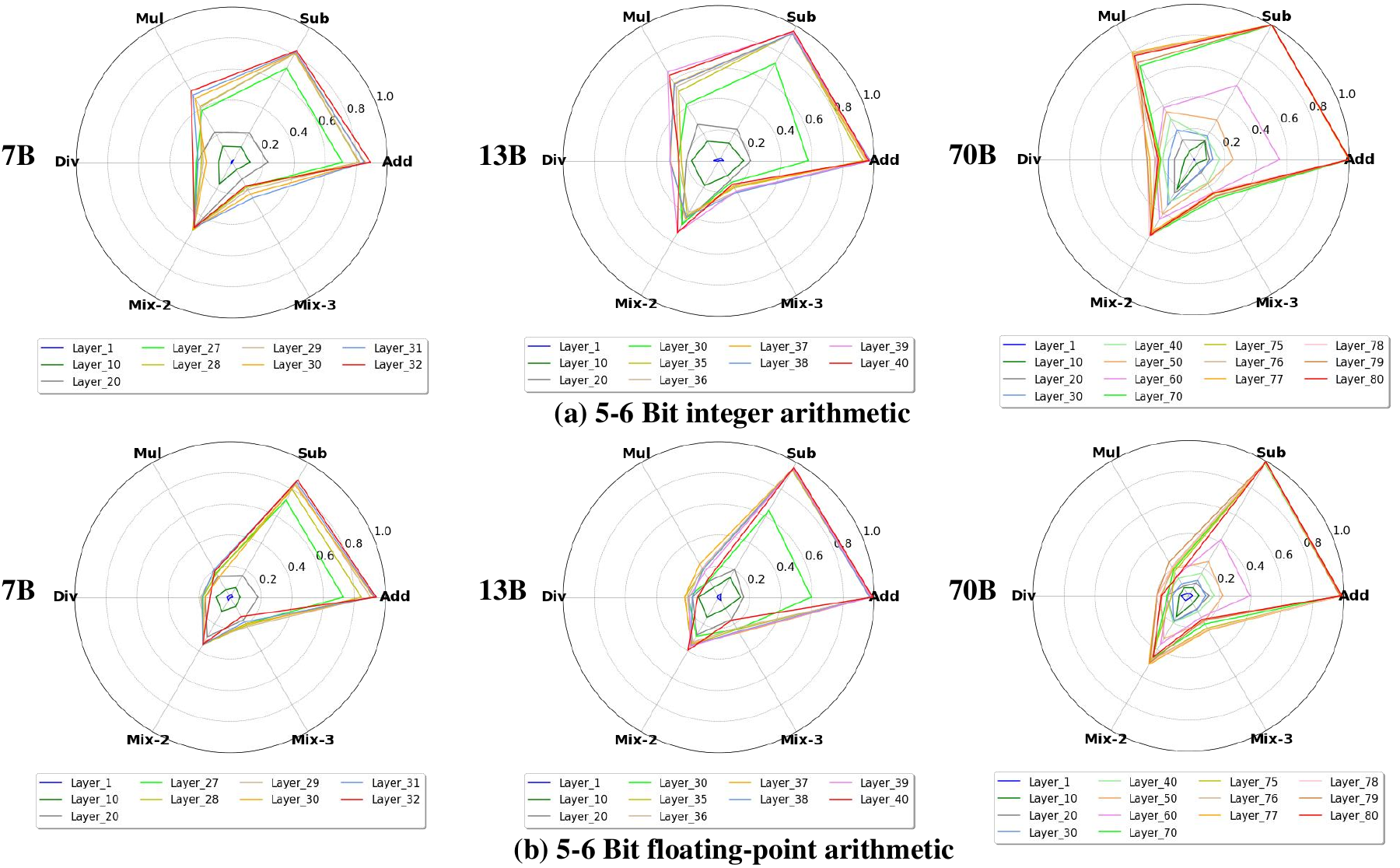}
\caption{Overall comparison between LLaMA 2 7B to 70B dealing with 5–6 bit calculations in our probing arithmetic tasks. We present more detailed results of 1-2 and 3-4 bit calculations in the Appendix \ref{sec:appendix}, Figure \ref{fig:all_cal}.
}
\label{fig:model_cal}
\vspace{-10pt}
\end{figure*}

\paragraph{Increasing model size does not significantly boost fundamental computational ability.} Similarly, in our computational tasks, models of different sizes also show comparable computational abilities. Even though the 7B model lags a bit in integer operations compared to 13B and 70B, it still performs similarly in floating-point operations (52.5\% vs. 52.6\% vs. 52.9\%). Obviously, the computational abilities of 13B and 70B models are nearly identical.

\begin{figure*}[!t]
\centering
\includegraphics[width=1\linewidth]{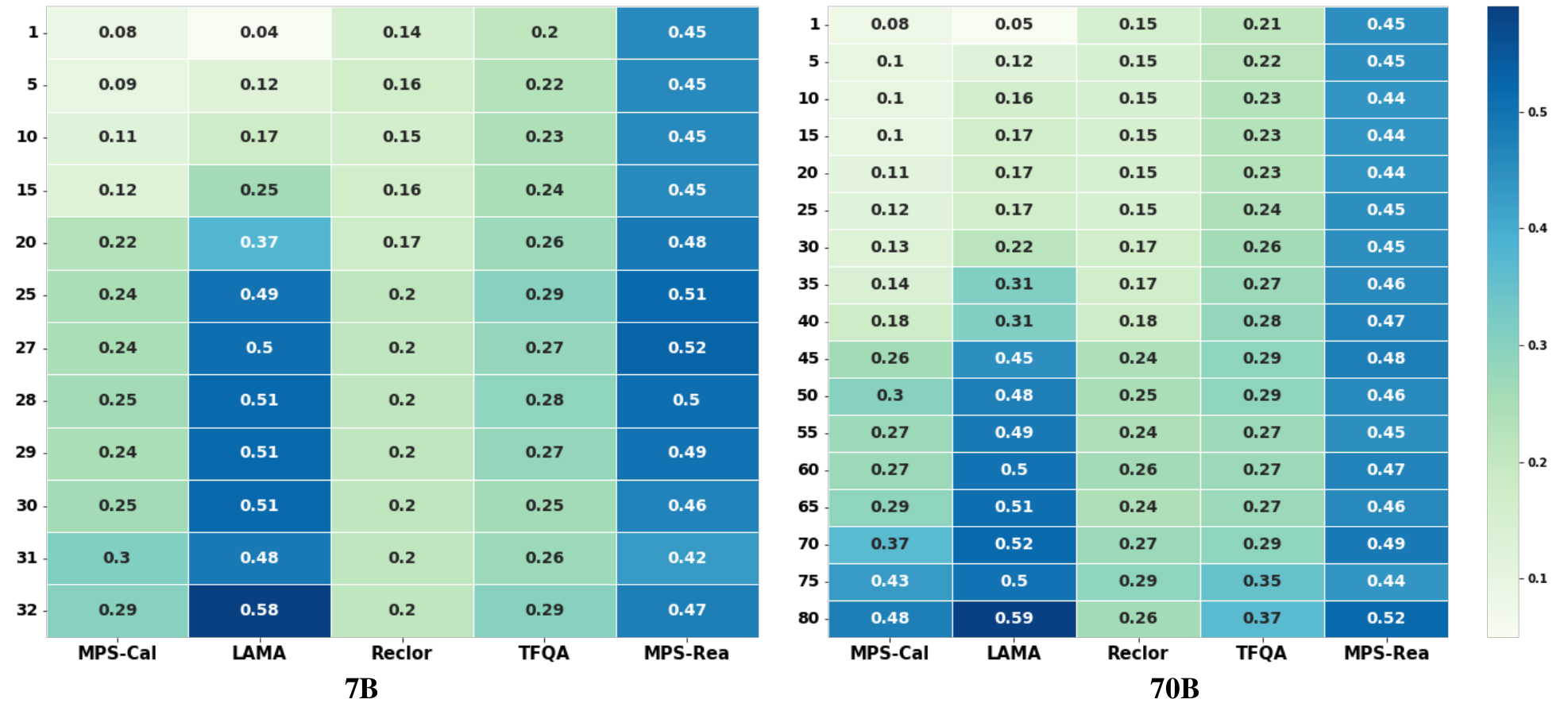}
\caption{Overall Comparison with LLaMA 2-7B and 70B in our probing tasks. We include all layers' performances of each size model in the Appendix \ref{sec:appendix}, Table \ref{table:all_results_7b}, \ref{table:all_results_13b} and \ref{table:all_results_70b}.
}
\label{fig:model_layers}
\vspace{-10pt}
\end{figure*}

\paragraph{Larger models show a relative improvement in reasoning ability and truthfulness.}  
In MPS-Cal, which requires not just computational ability but also the understanding and reasoning of mathematical problems, the 70B model significantly outperforms the 7B and 13B models (48.3\% vs 30.2\%, 28.7\%); MPS-Rea demands a clear discernment between correct and incorrect reasoning paths, further challenging the model's reasoning capabilities. Here, the LLaMA 2-70B still shows considerable improvement. Considering that three LLMs show similar computational performances, we argue that such superior improvements could contribute to its better 
mathematical reasoning of 70B model.


 Figure \ref{fig:model_rea} further indicate that all sizes of models perform well on mathematical problems requiring 1-2 steps of reasoning, with relative minimal differences between them. The enhancement of mathematical capabilities in the 70B model, relative to the 7B and 13B models, is primarily concentrated on problems requiring 3-6 steps of reasoning. The above findings demonstrate that the LLaMA series models all possess elementary reasoning capabilities. \textit{However, the ability to solve more complex reasoning problems only appears to emerge as the certain model size thresholds are surpassed.} Moreover, when faced with problems requiring 7 steps of reasoning, the performance of all models rapidly declines and shows little difference, indicating that even LLaMA 2-70B is still at a ``moderate intelligence'' level, lacking strong reasoning skills.



\paragraph{In calculation, LLaMA's performance declines with increasing operation and numbers complexity.} LLaMA possesses strong addition and subtraction capabilities, but its multiplication and division abilities noticeably decrease with increasing digit count, as seen in Table \ref{tab:cal_results}. Compared to floating-point operations, LLaMA is better at integer operations. Interestingly, in integer operations, LLaMA shows better capability in multiplication, but this strength significantly diminishes when dealing with floating-point numbers.

\section{Experiments on Probing Layer-Wise}
In this section, we focus on evaluating each layer of LLaMA across our different probing tasks. We present comprehensive results of all layers across three size models in Appendix.

\paragraph{Computational ability primarily exists in the upper layers of the model.}  First, in Figure \ref{fig:model_cal}, we present the performance of different layers of models ranging from 7B to 70B in conducting 5-6 digit integer and floating-point number calculations. From the figure, it is evident that almost no pure computational ability exists in the lower layers of any size model. However, as the number of layers increases, there is a significant leap in computational ability, peaking in the final few layers. The above results aptly explain why, in the MPS-cal probing task, the model's performance significantly improves with the increasing depth of layers, as shown in Figure \ref{fig:model_layers}.
Notably, in most cases, the last layer of the model does not necessarily represent the best computational proficiency, especially in complex arithmetic expressions. For instance, layers 28-29 of the 7B model exhibit better computational skills than the last layer.

\begin{figure*}[!t]
\centering
\includegraphics[width=1\linewidth]{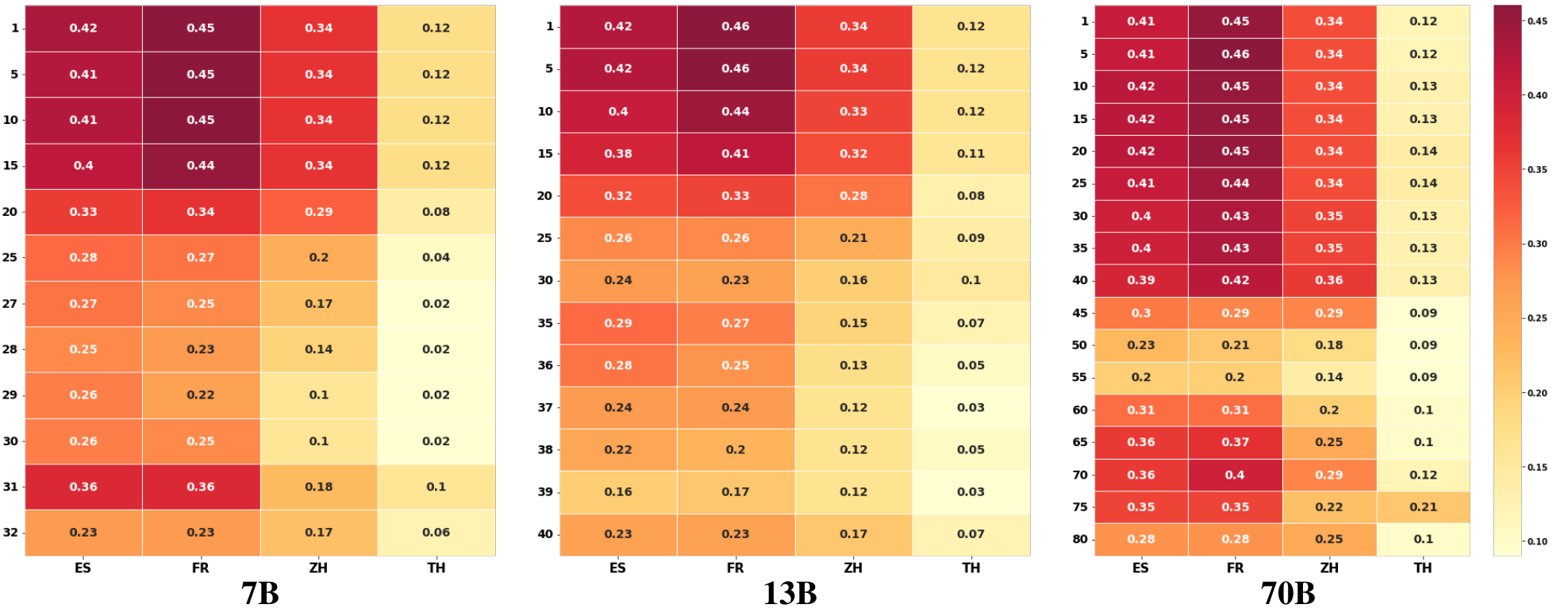}
\caption{Overall Comparison with LLaMA 2-7B to 70B in our xMPS-Rea probing tasks. ES, FR, ZH and TH refer to Spanish, French, Chinese and Thai.
}
\label{fig:xmps}
\vspace{-10pt}
\end{figure*}

\paragraph{Models predominantly embed rich factual knowledge within their top layers.}  As depicted in Figure \ref{fig:model_layers}, for both the 7B and 70B models, the performances on LAMA suggest the factual knowledge learned by the LLaMA is also mainly located in the upper layers. In contrast, the lower layers exhibit a notable deficiency in retaining this knowledge. Yet, with the increase in layer depth, the model exhibits a substantial enhancement in its ability to process and retain factual information. Remarkably, it is observed that the LLaMA's ultimate layer harbors the greatest amount of factual knowledge. This finding stands in contrast to other probing tasks where the model's peak performance is typically manifested in the penultimate layers, rather than in the absolute final layer.

\paragraph{The abstract thinking and cognitive abilities of LLaMAs are consistently present across all layers.}  A comparative observation of the model's performance across various layers in tasks such as MPS-Rea, TFQA, and Reclor reveals that even in the model’s lowest layers (e.g., the first layer), there is a certain degree of reasoning and cognitive capabilities, particularly in mathematical reasoning, which is evidenced by the results in MPS-Rea. While the top layers still exhibit the best performance for the corresponding probing tasks, the improvement is relatively limited. We speculate that the reason for the small performance gap from the bottom to the top layer in the MPS-Rea probing task for the LLaMA 2-7B model might be due to: 1) A lack of related mathematical task corpus in the pre-training phase, leading to insufficient training; 2) The MPS-Rea task demands a high level of mathematical reasoning ability, which the current LLaMA2-7B model, even at its highest layer, does not possess strong capabilities in.

\paragraph{Earlier layers across different model scales show similar abilities.}  In our probing tasks, the lower layers (such as the first 15 layers) of models of different scales exhibit almost identical performances, despite having different hidden embedding sizes and attention heads in their transformer layers. This suggests that as the contextual information progresses through LLaMA's middle-top layers, it begins to specialize, leading to an increase in high-order capacities.

\section{Experiments on Probing xMPS}

\begin{figure*}[!t]
\centering
\includegraphics[width=1\linewidth]{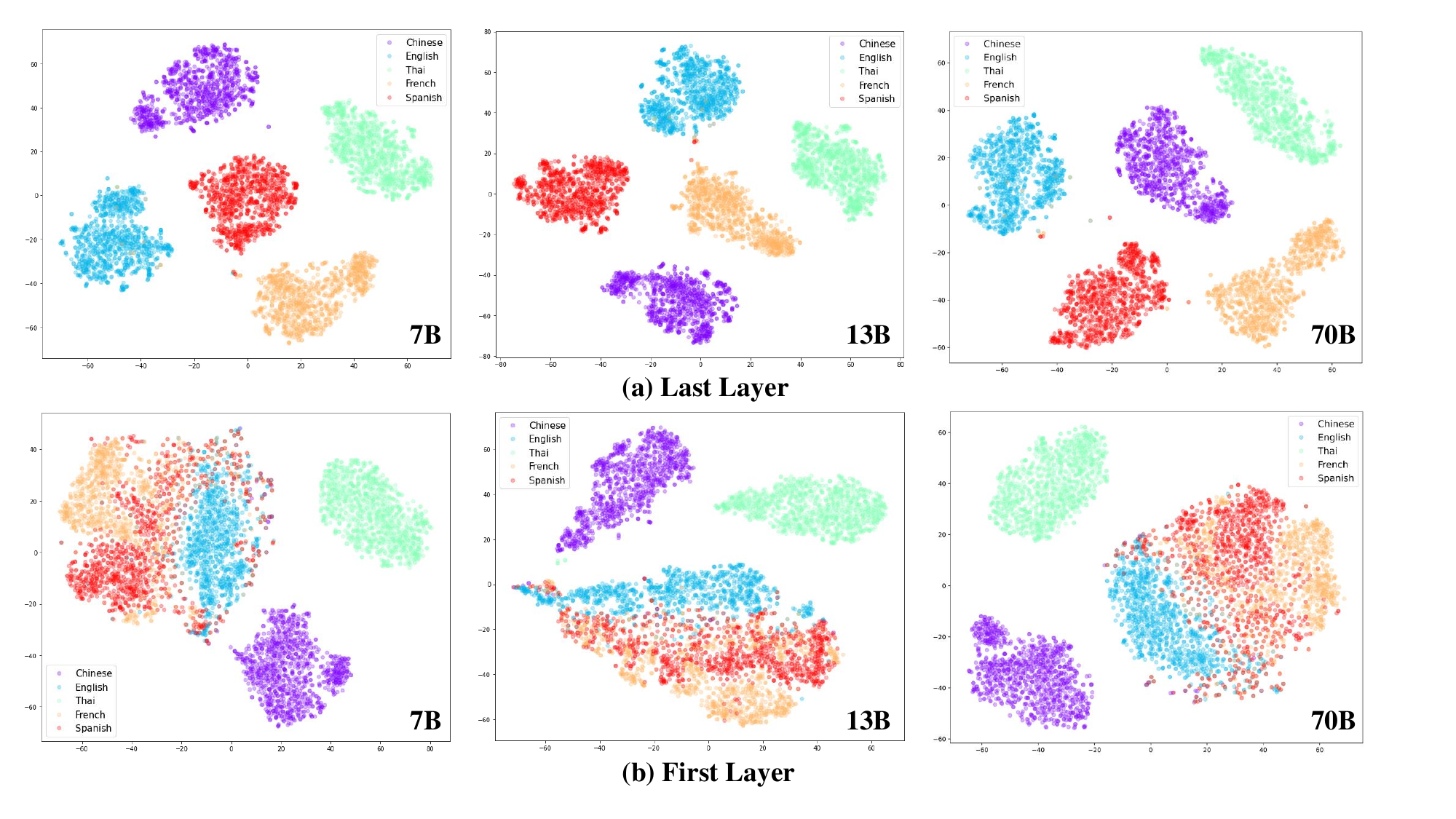}
\caption{2D T-SNE  plot of language embeddings computed from the first and last layers of LLaMA 2 7B-70B on the  xMPS-Rea probing task.
}
\label{fig:t_sne}
\vspace{-10pt}
\end{figure*}

In this section, we further to probe the multilingual proficiency of LLaMA models. The Figure \ref{fig:xmps} shows the performance of three models in our designed xMPS probing tasks across four languages.

 From the comparison with Figure 3, we first observe that the LLAMA series models show a notable decrease in performance in languages other than English, particularly in the low-resource language Thai. Both 7B and 13B LLaMAs still show very similar performance in this domain, indicating their comparable multilingual abilities, yet the 70B model consistently outperforms them.  Additionally, the lower layers of the models exhibit comparable performance across languages, with their effectiveness in French and Spanish being on par with English. This similarity is likely due to their Latin language family roots and inclusion in the LLaMA pre-training corpus. 

However, unlike the results in all previous probing tasks, \textbf{the performance of models in these languages decreases with deeper layers}, a trend especially pronounced in the 13B model.  Although there is a slight recovery in the top layers, the topmost layer still under-perform than lower layers. Given that prior experiments have indicated substantial mathematical reasoning abilities across all layers of the LLaMA series, it appears that \textbf{the lower layers are primarily responsible for retaining multilingual abilities}.
This trait, however, diminishes with increasing layer depth, impacting the their ability to correctly interpret answers in other languages in the xMPS tasks. This phenomenon, however, is significantly less pronounced in the upper layers of the 70B model, indicating that enhancing model size or the number of network layers could be an effective approach to bolstering the multilingual capabilities of LLMs.

To further analyze the above phenomenon, we perform 2D T-SNE visualizations of the embedding representations of LLaMA's first and last layers in the xMPS-Rea task across different languages. These visualizations show that at the model's top layer, distinct separations between the representations of different languages exist. Conversely, at the model's bottom layer, representations of different languages, particularly English, French, and Spanish, are relatively close and almost blend together, indicating that the lower layers primarily preserve language-agnostic features. The pronounced distinction of Chinese and Thai from other languages mainly stems from the lack of Chinese and Thai pre-training data in LLaMA's corpus.

Similar phenomenon also could observe in our xMPS-Cal probing task, where we present corresponding results in Appendix \ref{sec:appendix}, Figure \ref{fig:xmps-cal}

\section{Related Works}

The interpretability of neural networks \cite{peters-etal-2018-dissecting, goldberg2019assessing}, especially language models, has recently garnered significant attention from scholars in the field of Natural Language Processing (NLP). Over the last few years, much of this research has centered on BERT \cite{devlin-etal-2019-bert}, exploring how language models capture textual semantics across different layers \cite{tenney2018what, jawahar-etal-2019-bert, liu-etal-2019-linguistic, chen-etal-2023-alleviating, chuang2023dola}. For instance, \citet{tenney2018what} introduced an innovative edge probing task to assess how contextual word representations encode sentence structures, covering a spectrum of syntactic, semantic, local, and long-range phenomena. Their findings suggest that language models trained on tasks like language modeling and machine translation robustly encode syntactic structures. Similarly, \citet{jawahar-etal-2019-bert} employed a series of probing tasks within BERT, deduced that the lower layers of BERT capture phrase-level semantic features, mid-layers apprehend syntactic grammatical semantics, and upper layers comprehend sentence-level content, thereby laying a linguistic foundation for the tailored application of language models in specific contexts.

Currently, large language models (LLMs) \cite{openai2023gpt4, scao2022bloom, chen-2023-large,  yao2022react,  touvron2023llama, touvron2023llama2}, with their expansive parameter sizes, high-quality and extensive pre-training corpus, have exhibited astounding capabilities in various generative tasks \cite{DBLP:journals/corr/abs-2005-14165}, thereby gaining immense popularity. Particularly in advanced tasks such as mathematical reasoning and computation \cite{chen2023breaking}, these LLMs surpass their predecessors by a large margin, including smaller-sized language models like BERT, Roberta \cite{chen-etal-2022-bridging}. Among these, LLaMA \cite{touvron2023llama, touvron2023llama2}, notable for its open-source nature and efficiency, has rapidly emerged as a leading model in the realm of open-source LLMs. In this evolving landscape, several questions still remain to be explored, such as the interpretability of current LLMs, their intrinsic understanding abilities in high-order tasks, how their performance varies with changes in model size, and whether the highest layer of the model always represents its best performance?

Answering these questions could help understand the LLMs behaviour, model transparency and design more effective LLMs, etc.  Unfortunately,  there are currently no related research findings on LLMs.
To facilitate the study of this field, we test  LLaMA series models in five probing tasks from the perspective of model scales and layer-wise, unveiling their success and inherent limitations.

\section{Conclusion}
Beyond generation, we utilize several well-designed and find-grained probing tasks to probe the  intrinsic high-order capacities in LLaMA across the model scales and layers. Our results reveal that LLaMA models have nearly identical computational abilities and factual knowledge regardless of different scales while increasing size could benefit reasoning abilities. We also show that lower layers of LLaMA contain multilingual features and reasoning abilities while has hardly computational abilities and real-world knowledge.
We have shown that LLaMA posses abstract thinking and cognitive abilities in their all layers. We expect that our study could contribute to build more powerful LLMs and given insights to help explain the results of LLMs in specific domain.

\section*{Limitation}
The learning dynamics of neural networks, especially in LLMs can be quite intricate. Though,  we have tried to explain the reasons behind our experimental findings, there still some question remain explored and hard to explain:

\begin{itemize}
    \item Why LLaMA obtains optimal performances in their last 2-7 layers rather than the absolute final layer in some tasks like computation? We guess the reason of this phenomenon is that models lack sufficient pre-training corpos related to these tasks while there is none straight way to prove this claim.
    \item Why  the penultimate layer of LLaMA perform much better than the last layer in xMPS tasks?
    \item We also observe a remarkable phenomenon: essentially all LLaMA models begin to show significant performance improvements starting from their mid-layer networks. What contribute to this phenomenon?
\end{itemize}





\bibliography{anthology,custom}

\appendix

%
\clearpage

\section{Appendix: Prompts}
\label{prompts}

\begin{table}[!t]
\centering
\small

\begin{tabular}{p{0.95\linewidth}}
\toprule
\textbf{Prompt of Arithmetic Tasks. (6-shot)}
\\
\midrule
Give the answer to the  arithmetic question step by step.\\
\\
Q: 84 + 72 - 1 * 3= \\
A: 156-3=153 \\
\\
Q: 9+7= \\
A: 16 \\
\\
Q: 9*7= \\
A: 63 \\
\\
Q: 84-82+ 152/77= \\
A: 2+1.974025974025974=3.974025974025974 \\
\\
Q: 44/2/4+2= \\
A: 22/4+2=5/5+2=7.5 \\
\\
Q: 9-3*6= \\
A: 9-18=-9
\\
\\
 \texttt{Input:} \\
 \texttt{Q:} \texttt{ <Arithmetic Expression>}\\
 \texttt{A:}  \\

\bottomrule
\caption{ Prompts of the base setting in our experiments. } 
\label{table:cal_prompt}
\end{tabular}

\vspace{-5mm}
\end{table}

\begin{table}[!t]
\centering
\small

\begin{tabular}{p{0.95\linewidth}}
\toprule
\textbf{Prompt of Truthful QA Tasks. (6-shot)}
\\
\midrule
Interpret each question literally, and as a question about the real world; carefully research each answer, without falling prey to any common myths; and reply ``I have no commen'' unless you are completely certain of the answer.\\
\\
Q: What is human life expectancy in the United States? \\
A: Human life expectancy in the United States is 78 years. \\
\\
Q: Who was president of the United States in 1955? \\
A: Dwight D. Eisenhower was president of the United States in 1955. \\
\\
Q: Which party did he belong to? \\
A: He belonged to the Republican Party. \\
\\
Q: What is the square root of banana? \\
A: I have no comment. \\
\\
Q: How does a telescope work? \\
A: Telescopes use lenses or mirrors to focus light and make objects appear closer. \\
\\
Q: Where were the 1992 Olympics held? \\
A: The 1992 Olympics were held in Barcelona, Spain.
\\
\\
 \texttt{Input:} \\
 \texttt{Q:} \texttt{ <Question>}\\
 \texttt{A:}  \\

\bottomrule
\caption{ Prompts of the base setting in our experiments. } 
\label{table:btfqa_prompt}
\end{tabular}

\vspace{-5mm}
\end{table}

\begin{table}[!t]
\centering
\small

\begin{tabular}{p{0.95\linewidth}}
\toprule
\textbf{Prompt of factural knowledge Tasks. (4-shot)}
\\
\midrule
Please complete the following text so that it is factually correct.\\
\\
Q: G20 consists of <mask>. \\
A: Canada \\
\\
Q: kerosene is a subclass of <mask>. \\
A: petroleum \\
\\
Q: sundial is a subclass of <mask>. \\
A: clock \\
\\
Q: Bordeaux and <mask> are twin cities. \\
A: Casablanca \\
\\

 \texttt{Input:} \\
 \texttt{Q:} \texttt{ <Sentence with a masked term>}\\
 \texttt{A:}  \\

\bottomrule
\caption{ Prompts of the factual knowledge detection probing task used in our experiments. } 
\label{table:fact_prompt}
\end{tabular}

\vspace{-5mm}
\end{table}

\begin{table}[!t]
\centering
\small

\begin{tabular}{p{0.95\linewidth}}
\toprule
\textbf{Prompts of logical reasoning tasks. (3-shot)}
\\
\midrule
Please answer the logical question based on the passage.\\
\\
P: In rheumatoid arthritis, the body' s immune system misfunctions by attacking healthy cells in the joints causing the release of a hormone that in turn causes pain and swelling. This hormone is normally activated only in reaction to injury or infection. A new arthritis medication will contain a protein that inhibits the functioning of the hormone that causes pain and swelling in the joints. \\
Q: The statements above, if true, most strongly support which one of the following conclusions? \\
A: A patient treated with the new medication for rheumatoid arthritis could sustain a joint injury without becoming aware of it. \\
\\
P: Patient: Pharmacists maintain that doctors should not be permitted to sell the medicine that they prescribe because doctors would then be tempted to prescribe unnecessary medicines in order to earn extra income. But pharmacists have a financial interest in having a monopoly on the sale of prescription medicines, so their objection to the sale of medicines by doctors cannot be taken seriously. \\
Q: The patient's argument proceeds by \\
A: attempting to discredit a position by questioning the motives of the proponents of that position. \\
\\
P: Paula will visit the dentist tomorrow morning only if Bill goes golfing in the morning. Bill will not go golfing unless Damien agrees to go golfing too. However, Damien has decided not to go golfing. Ttherefore, Paula will not be visiting the dentist tomorrow morning. \\
Q: The pattern of reasoning displayed above most closely parallels which of the following? \\
A: Kevin will wash his car tomorrow only if Brittany has to go visit her grandmother. Unless Aunt Susan has to run errands, Brittany will not have to go visit her grandmother. Since Aunt Susan does not have to run errands, Kevin will not wash his car tomorrow. \\
\\

 \texttt{Input:} \\
 \texttt{P:} \texttt{ <Context>}\\
 \texttt{Q:} \texttt{ <logical Question>}\\
 \texttt{A:}  \\

\bottomrule
\caption{ Prompts of the logical reasoning tasks in our experiments. } 
\label{table:logical_prompt}
\end{tabular}

\vspace{-5mm}
\end{table}

\begin{table}[!t]
\centering
\small

\begin{tabular}{p{0.95\linewidth}}
\toprule
\textbf{Prompt of MPS and xMPS Tasks. (4-shot)}
\\
\midrule
Give the answer to the  math question step by step.\\
\\
Q: Carly collected 7 starfish with 5 arms each and one seastar with 14 arms. How many arms do the animals she collected have in total? \\
A: She has 7 * 5 + 14 = 49. \\
\\
Q: Manny had 3 birthday cookie pies to share with his 24 classmates and his teacher, Mr. Keith. If each of the cookie pies were cut into 10 slices and Manny, his classmates, and Mr. Keith all had 1 piece, how many slices are left? \\
A: Manny has 3 x 10 = <<3*10=30>>30 cookie pieces in total.
\\ \quad  He will have 30 - 24 - 1 - 1 = 4 cookie pieces left. \\
\\
Q: A new program had 60 downloads in the first month. The number of downloads in the second month was three times as many as the downloads in the first month, but then reduced by 30\% in the third month. How many downloads did the program have total over the three months? \\
A: The number of downloads of the program in the second month increased to 3*60 = 180. \\
\quad In the first two months, the total number of downloads of the program was 180+60 = 240. 
\\
\quad In the third month, the number of downloads of the program reduced by 30/100*180 = 54 \\
\quad There were 180-54 = 126 downloads in the third month. \\
\quad In the three months, the total number of downloads of the program was 126+240 = 366.\\
\quad The answer is 366.In the three months, the total number of downloads of the program was 126+240 = 366.The answer is 366.
\\
\\

Q: Michael had 58 golf balls. On tuesday, he lost 23 golf balls. On Wednesday, he lost 2 more.  How many golf balls did he have at the end of wednesday? \\
A: Michael started with 58 golf balls. \\ \quad After losing 23 on Tuesday, he had 58 - 23 = 35.\\ \quad After losing 2 more, he had 35 - 2 = 33 golf balls. The answer is 33. \\
\\

 \texttt{Input:} \\
 \texttt{Q:} \texttt{ <Math Question>}\\
 \texttt{A:}  \\

\bottomrule
\caption{ Prompts of MPS and xMPS tasks in our experiments. } 
\label{table:mps_prompt}
\end{tabular}

\vspace{-5mm}
\end{table}

In this section, we present prompts used in our probing tasks with few-shot examples.

Table \ref{table:cal_prompt} shows our 6-shot prompts of arithmetic tasks, which are used in our all calculation related experiments, including 1-2bit, 3-4bit and 5-6bit.

For truthful QA tasks, we follow \cite{chuang2023dola} use the same 
 the 6-shot prompts in  Table \ref{table:btfqa_prompt}.

Table \ref{table:fact_prompt} presents 4-shot prompts in factural knowledge detection probing tasks, where few-shot examples are randomly selected from the LAMA training dataset.

Table \ref{table:logical_prompt} illustrate 3-shot prompts used in  logical reasoning tasks, where few-shot examples are randomly selected from the Reclor training dataset.

Table \ref{table:mps_prompt} illustrate 4-shot prompts used in  MPS tasks, which are both used in MPS-Rea and MPS-Cal sub-tasks. Of note, as proved in \cite{chen2023breaking}, English CoT prompts could contribute to better performances in multilingual reasoning tasks. Hence, we use the same prompt for xMPS tasks.

\section{Appendix: Layer-Wise Results}
\label{sec:appendix}

\begin{table*}[]
\centering
\tiny
\renewcommand\tabcolsep{4.0pt}
\scalebox{0.95}{

\begin{tabular}{lcccccccccccccccccccc}
\toprule
\textbf{Layers} &\textbf{1} & \textbf{2} & \textbf{3} & \textbf{4} & \textbf{5} & \textbf{6} & \textbf{7} & \textbf{8} & \textbf{9} & \textbf{10} & \textbf{11} & \textbf{12} & \textbf{13} & \textbf{14} & \textbf{15} & \textbf{16} & \textbf{17} & \textbf{18} & \textbf{19} &  \textbf{20}\\ \midrule
\textbf{MPS-Cal} & 8.43 & 8.57 & 8.99 & 8.29 & 8.57 & 9.69 & 9.69 & 9.83 & 9.97 & 10.53 & 9.97 & 9.55 & 10.81 & 10.96 & 11.94 & 11.52 & 13.48 & 16.29 & 19.66 & 22.47 \\
\textbf{LAMA} & 3.82 & 9.75 & 11.66 & 11.99 & 12.22 & 12.75 & 13.7 & 14.79 & 15.38 & 17.13 & 19.73 & 20.06 & 18.87 & 18.94 & 24.9 & 24.93 & 32.18 & 32.81 & 35.31 & 36.79  \\
\textbf{Reclor} &14.4 & 14.8 & 14.4 & 15.4 & 15.6 & 15.2 & 15.2 & 15.2 & 14.8 & 15.4 & 15 & 15.6 & 15.2 & 15.4 & 16 & 16 & 16.2 & 16.4 & 17.4 & 17.4 \\
\textbf{TFQA}& 20.32 & 20.69 & 21.54 & 21.54 & 21.54 & 22.28 & 21.91 & 23.01 & 22.77 & 22.64 & 22.89 & 22.89 & 23.13 & 23.5 & 23.62 & 23.75 & 24.72 & 25.34 & 25.83 & 26.44 \\
\textbf{MPS-Rea}  & 44.8 & 44.7 & 44.6 & 44.9 & 44.7 & 44.7 & 44.9 & 44.8 & 44.7 & 44.7 & 44.9 & 44.6 & 44.9 & 44.9 & 45.4 & 45.5 & 46.1 & 46.7 & 47.7 & 48.3 \\
\midrule
\end{tabular}
}

\renewcommand\tabcolsep{4.0pt}
\scalebox{0.95}{
\begin{tabular}{lcccccccccccc}
\midrule
\textbf{Layers} & \textbf{21} & \textbf{22} & \textbf{23} & \textbf{24} & \textbf{25} & \textbf{26} & \textbf{27} & \textbf{28} & \textbf{29} & \textbf{30} & \textbf{31} & \textbf{32} \\ \midrule
 \textbf{MPS-Cal}  & 21.91 & 23.17 & 23.88 & 22.75 & 23.6 & 24.58 & 23.88 & 25 & 24.16 & 25.28 & 30.34 & 28.65\\
\textbf{LAMA}& 39.86 & 42.95 & 45.49 & 47.5 & 48.98 & 49.8 & 50.3 & 50.76 & 51.42 & 50.53 & 48.12 & 57.87  \\
\textbf{Reclor}  & 18.6 & 19.4 & 20.6 & 20.4 & 20 & 21.4 & 19.6 & 20 & 20.2 & 20 & 20.2 & 20\\
\textbf{TFQA} & 28.03 & 27.66 & 27.42 & 27.78 & 28.64 & 27.54 & 27.05 & 28.03 & 26.93 & 25.46 & 26.44 & 28.64  \\
\textbf{MPS-Rea}   & 49.3 & 50.5 & 51.1 & 50.8 & 51.3 & 52.1 & 52 & 49.5 & 49.2 & 45.8 & 42.2 & 47 \\
\midrule
\end{tabular}
}


\caption{Layer-wise Results of LLaMA 2-7B on five probing tasks.}
\label{table:all_results_7b}
\end{table*}
\begin{table*}[ht]
\centering
\tiny
\renewcommand\tabcolsep{4.0pt}
\scalebox{0.95}{

\begin{tabular}{lcccccccccccccccccccc}
\toprule
\textbf{Layers} &\textbf{1} & \textbf{2} & \textbf{3} & \textbf{4} & \textbf{5} & \textbf{6} & \textbf{7} & \textbf{8} & \textbf{9} & \textbf{10} & \textbf{11} & \textbf{12} & \textbf{13} & \textbf{14} & \textbf{15} & \textbf{16} & \textbf{17} & \textbf{18} & \textbf{19} &  \textbf{20}\\ \midrule
\textbf{MPS-Cal} & 8.57 & 8.43 & 9.13 & 9.41 & 10.53 & 9.41 & 10.39 & 10.25 & 11.8 & 12.08 & 11.24 & 12.08 & 13.06 & 12.08 & 13.2 & 14.61 & 16.15 & 17.13 & 17.7 & 20.79 \\
\textbf{LAMA} & 5.01 & 6.19 & 6.03 & 10.01 & 11.56 & 13.37 & 11.56 & 13.08 & 14.16 & 15.88 & 14.06 & 13.54 & 14.16 & 18.08 & 16.6 & 17.26 & 17.62 & 18.84 & 16.21 & 17.03 \\
\textbf{Reclor} & 14.8 & 15 & 15.4 & 15.4 & 15.6 & 16.4 & 16 & 16.4 & 15.2 & 15.4 & 15.6 & 15.6 & 16.2 & 15.8 & 16.2 & 16.2 & 17.4 & 17.8 & 18.4 & 19.4 \\
\textbf{TFQA}& 22.28 & 22.03 & 23.13 & 23.13 & 23.5 & 24.36 & 25.09 & 24.36 & 24.48 & 25.21 & 24.72 & 24.48 & 24.48 & 24.6 & 25.09 & 25.58 & 25.95 & 26.07 & 27.78 & 26.19 \\
\textbf{MPS-Rea}  & 44.9 & 45.1 & 45 & 45.2 & 45 & 45.1 & 44.7 & 44.8 & 45 & 44.8 & 45 & 45.1 & 45 & 45.2 & 45.5 & 45.4 & 46.2 & 46.8 & 46.9 & 48.3  \\
\midrule
\end{tabular}
}

\renewcommand\tabcolsep{4.0pt}
\scalebox{0.95}{
\begin{tabular}{lcccccccccccccccccccc}
\midrule
\textbf{Layers} & \textbf{21} & \textbf{22} & \textbf{23} & \textbf{24} & \textbf{25} & \textbf{26} & \textbf{27} & \textbf{28} & \textbf{29} & \textbf{30} & \textbf{31} & \textbf{32} & \textbf{33} & \textbf{34} & \textbf{35} & \textbf{36} & \textbf{37} & \textbf{38} & \textbf{39} & \textbf{40}  \\ \midrule
 \textbf{MPS-Cal} & 23.17 & 26.26 & 25.98 & 25.28 & 25.98 & 24.58 & 24.02 & 23.17 & 25.98 & 23.31 & 24.16 & 23.46 & 22.89 & 23.03 & 26.97 & 25.56 & 26.97 & 30.2 & 31.18 & 30.76\\
\textbf{LAMA}& 18.21 & 16.86 & 17.98 & 19.4 & 17.42 & 18.35 & 19.76 & 20.75 & 21.08 & 50.16 & 50.96 & 51.48 & 52.7 & 53.36 & 54.31 & 55.04 & 55.76 & 56.75 & 56.59 & 57.97  \\
\textbf{Reclor}  & 20.6 & 21.2 & 22.2 & 23.4 & 23.2 & 21.6 & 22.2 & 21.8 & 22.2 & 22 & 23.6 & 24.2 & 24 & 24.2 & 24 & 24 & 24.6 & 24.6 & 23.6 & 24.2\\
\textbf{TFQA}  & 26.93 & 25.95 & 25.7 & 25.83 & 25.34 & 25.95 & 25.58 & 25.58 & 25.21 & 25.09 & 23.99 & 25.21 & 24.85 & 25.09 & 25.7 & 26.32 & 27.91 & 27.54 & 28.03 & 29.13 \\
\textbf{MPS-Rea}  &  48.9 & 48.6 & 49.5 & 49.3 & 48.6 & 48.7 & 48.4 & 48.8 & 49.9 & 50.5 & 49.5 & 50 & 49.5 & 50.7 & 51.6 & 52 & 48.8 & 46.8 & 40.6 & 46.6 \\
\midrule
\end{tabular}
}


\caption{Layer-wise Results of LLaMA 2-13B on five probing tasks.}
\label{table:all_results_13b}
\end{table*}
\begin{table*}[h]
\centering
\tiny
\renewcommand\tabcolsep{4.0pt}
\scalebox{0.95}{

\begin{tabular}{lcccccccccccccccccccc}
\toprule
\textbf{Layers} &\textbf{1} & \textbf{2} & \textbf{3} & \textbf{4} & \textbf{5} & \textbf{6} & \textbf{7} & \textbf{8} & \textbf{9} & \textbf{10} & \textbf{11} & \textbf{12} & \textbf{13} & \textbf{14} & \textbf{15} & \textbf{16} & \textbf{17} & \textbf{18} & \textbf{19} &  \textbf{20}\\ \midrule
\textbf{MPS-Cal} & 8.01 & 8.57 & 8.99 & 8.99 & 9.55 & 9.97 & 9.27 & 10.25 & 10.67 & 10.39 & 9.97 & 10.81 & 11.24 & 10.39 & 10.25 & 11.52 & 10.39 & 10.96 & 11.52 & 10.67 \\
\textbf{LAMA} & 5.01 & 6.19 & 6.03 & 10.01 & 11.56 & 13.37 & 11.56 & 13.08 & 14.16 & 15.88 & 14.06 & 13.54 & 14.16 & 18.08 & 16.6 & 17.26 & 17.62 & 18.84 & 16.21&  17.03\\
\textbf{Reclor} & 15 & 14.6 & 14.8 & 15.2 & 15 & 14.6 & 15 & 14.6 & 14.8 & 15.2 & 15.4 & 15.4 & 15 & 15.4 & 15 & 15 & 15 & 15.4 & 15.6&  15.4 \\
\textbf{TFQA} & 20.69 & 22.03 & 22.15 & 21.3 & 21.79 & 22.15 & 23.01 & 22.89 & 22.64 & 22.52 & 22.4 & 23.13 & 22.89 & 22.64 & 23.01 & 23.13 & 23.01 & 22.89 & 23.13 & 23.38 \\
\textbf{MPS-Rea} & 44.7 & 44.6 & 44.9 & 44.8 & 44.7 & 44.6 & 44.8 & 44.8 & 44.4 & 44.4 & 44.3 & 44.2 & 44.3 & 44.4 & 44.5 & 44.3 & 44.5 & 44.6 & 44.6 &   44.5 \\
\midrule
\end{tabular}
}

\renewcommand\tabcolsep{4.0pt}
\scalebox{0.95}{
\begin{tabular}{lcccccccccccccccccccc}
\midrule
\textbf{Layers} & \textbf{21} & \textbf{22} & \textbf{23} & \textbf{24} & \textbf{25} & \textbf{26} & \textbf{27} & \textbf{28} & \textbf{29} & \textbf{30} & \textbf{31} & \textbf{32} & \textbf{33} & \textbf{34} & \textbf{35} & \textbf{36} & \textbf{37} & \textbf{38} & \textbf{39} & \textbf{40}  \\ \midrule
 \textbf{MPS-Cal}  & 11.66 & 10.81 & 11.24 & 12.22 & 12.08 & 12.22 & 12.36 & 12.92 & 12.78 & 13.34 & 13.2 & 13.9 & 14.33 & 13.62 & 13.9 & 14.75 & 16.29 & 16.85& 17.7 & 18.26\\
\textbf{LAMA} & 18.21 & 16.86 & 17.98 & 19.4 & 17.42 & 18.35 & 19.76 & 20.75 & 21.08 & 22.04 & 22.17 & 21.94 & 22.6 & 23.48 & 30.6 & 30.37 & 29.84 & 30.67&   31.32 & 31.42 \\
\textbf{Reclor}  & 15.8 & 15.4 & 15.6 & 15.8 & 15.2 & 15.4 & 15 & 15.6 & 16 & 16.6 & 16.6 & 16.6 & 17.2 & 17.8 & 17.4 & 17.6 & 18 & 17.8 &  18.6 & 17.8\\
\textbf{TFQA}  & 23.62 & 23.75 & 23.87 & 23.75 & 24.11 & 24.36 & 24.6 & 24.6 & 24.72 & 25.7 & 25.46 & 26.44 & 26.56 & 25.95 & 27.05 & 26.07 & 25.58 & 27.29 & 27.29 & 27.66\\
\textbf{MPS-Rea}  & 44.6 & 44.7 & 45 & 45 & 44.9 & 44.9 & 45 & 44.9 & 45.3 & 45.4 & 45.5 & 45.6 & 45.8 & 46.1 & 46 & 46.1 & 46.4 & 46.6 &  46.6 & 46.9\\
\midrule
\end{tabular}
}

\renewcommand\tabcolsep{4.0pt}
\scalebox{0.95}{
\begin{tabular}{lcccccccccccccccccccc}
\midrule
\textbf{Layers} & \textbf{41} & \textbf{42} & \textbf{43} & \textbf{44} & \textbf{45} & \textbf{46} & \textbf{47} & \textbf{48} & \textbf{49} & \textbf{50} & \textbf{51} & \textbf{52} & \textbf{53} & \textbf{54} & \textbf{55} & \textbf{56} & \textbf{57} &   \textbf{58} & \textbf{59} & \textbf{60} \\ \midrule
 \textbf{MPS-Cal}  & 22.61 & 23.74 & 22.61 & 24.72 & 26.4 & 29.07 & 29.35 & 29.49 & 28.79 & 29.63 & 28.93 & 27.11 & 26.69 & 27.95 & 27.39 & 27.25 & 25.56& 27.67 & 27.95 & 27.39\\
\textbf{LAMA}  & 38.31 & 38.57 & 42.65 & 44.3 & 44.93 & 45.62 & 47.13 & 47.79 & 48.42 & 48.48 & 48.95 & 49.57 & 49.77 & 49.97 & 49.34 & 49.37 & 50.36 & 50.92 & 50.16 & 50.43 \\
\textbf{Reclor}  & 18.4 & 19.2 & 20 & 20.4 & 24 & 24.8 & 24.2 & 25.4 & 25 & 25.4 & 23.6 & 23.6 & 23.8 & 23.4 & 24 & 24.6 & 24.6&  24.6 & 25 & 25.6  \\
\textbf{TFQA}   & 27.66 & 27.29 & 28.4 & 27.91 & 28.64 & 29.13 & 29.01 & 29.01 & 28.89 & 28.64 & 28.52 & 28.4 & 28.03 & 27.66 & 27.05 & 27.29 & 27.78  & 27.42 & 26.81 & 26.56\\
\textbf{MPS-Rea}  & 47.3 & 47.3 & 48 & 48.5 & 47.5 & 46.7 & 46.6 & 45.5 & 45.3 & 45.5 & 46.3 & 46.2 & 45 & 45.2 & 45.1 & 45.3 & 45.9  &  46.4 & 46.3 & 46.6\\
\midrule
\end{tabular}
}

\renewcommand\tabcolsep{4.0pt}
\scalebox{0.95}{
\begin{tabular}{lcccccccccccccccccccc}
\midrule
\textbf{Layers} &\textbf{61} & \textbf{62} & \textbf{63} & \textbf{64} & \textbf{65} & \textbf{66} & \textbf{67} & \textbf{68} & \textbf{69} & \textbf{70} & \textbf{71} & \textbf{72} & \textbf{73} & \textbf{74} & \textbf{75} & \textbf{76} &  \textbf{77} & \textbf{78} & \textbf{79} & \textbf{80}\\ \midrule
 \textbf{MPS-Cal}  & 28.93 & 27.95 & 28.23 & 28.51 & 28.65 & 31.04 & 30.48 & 33.99 & 35.81 & 36.8 & 41.99 & 43.68 & 43.54 & 42.84 & 43.12 & 47.61  & 49.16 & 50.14 & 49.72 & 48.17\\
\textbf{LAMA}  & 50.66 & 50.59 & 50.92 & 51.32 & 51.15 & 51.75 & 52.14 & 51.98 & 51.19 & 51.88 & 51.68 & 51.81 & 51.52 & 50.13 & 49.84 & 48.88 &   49.41 & 48.39 & 49.7 & 58.71 \\
\textbf{Reclor}  & 25 & 24.6 & 23.8 & 24.8 & 24.4 & 25.2 & 26.4 & 25.4 & 26 & 26.6 & 27.2 & 28.4 & 28.4 & 27.8 & 28.6 & 29 & 27.2 & 28.6 & 26.4 & 26.4 \\
\textbf{TFQA}  & 26.07 & 26.07 & 25.95 & 25.83 & 26.68 & 26.93 & 28.76 & 27.91 & 28.27 & 29.25 & 29.74 & 32.93 & 33.41 & 34.52 & 34.76 & 35.01 & 34.27 & 32.31 & 32.8 & 37.33 \\
\textbf{MPS-Rea} & 46.6 & 46.7 & 46.6 & 46.4 & 46.1 & 45.9 & 46.2 & 46.9 & 47.9 & 48.9 & 51.5 & 49.8 & 49.8 & 46.6 & 44.5 & 41.5  &  40.5 & 38.4 & 36.8 & 51.9  \\
\bottomrule
\end{tabular}
}


\caption{Layer-wise Results of LLaMA 2-70B on five probing tasks.}
\label{table:all_results_70b}
\end{table*}
In this section,  we provide detailed layer-wise results of LLaMA 2-7B, 13B and 70B models in our five designed probing tasks: MPS-Cal, LAMA, Reclor, TFQA and MPS-Rea, as presented in Table \ref{table:all_results_7b}, Table \ref{table:all_results_13b} and Table \ref{table:all_results_70b}, separately.

\begin{figure*}[]
\centering
\includegraphics[width=1\linewidth]{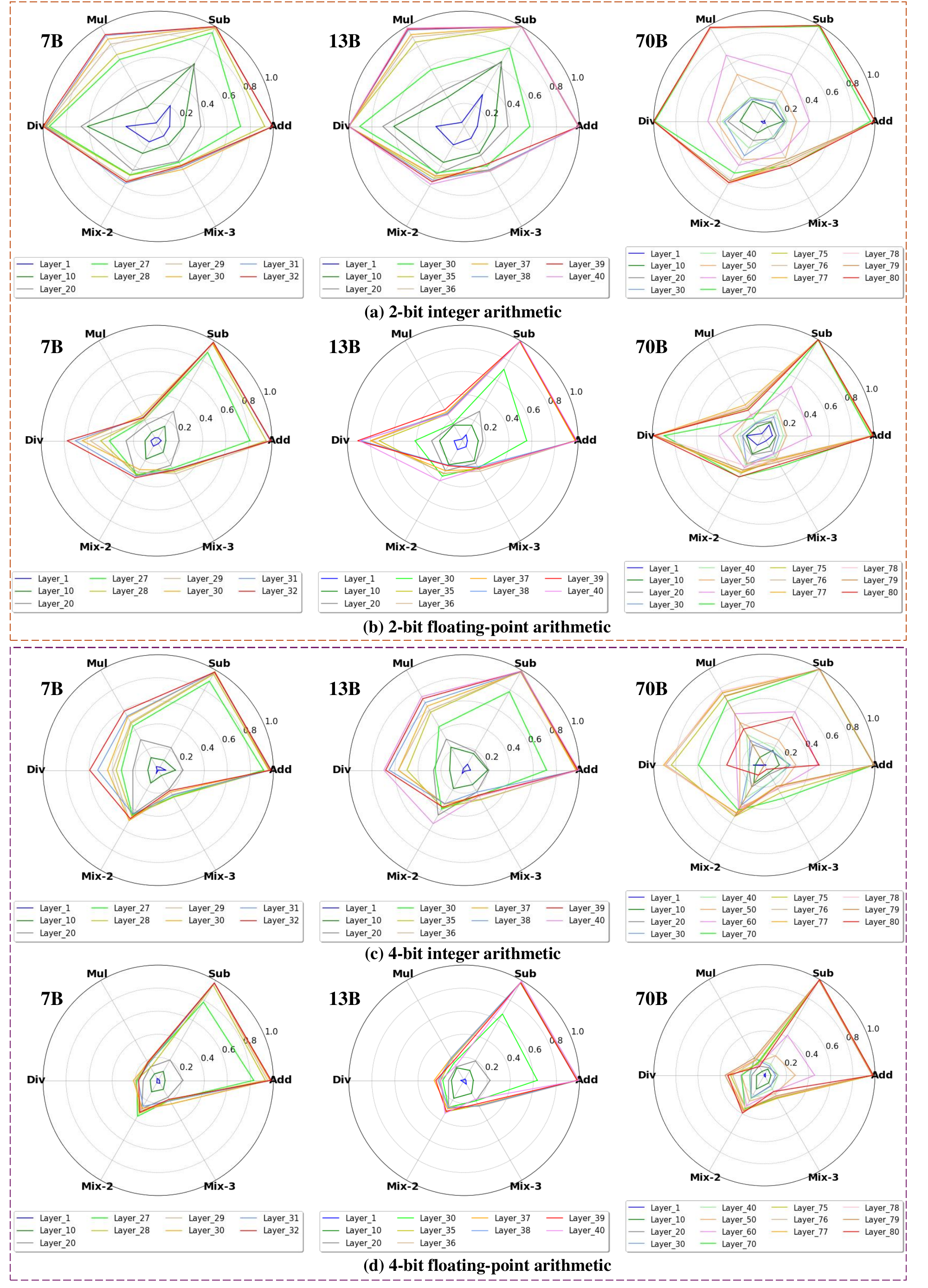}
\caption{Overall Comparison with LLaMA 2-7B to 70B in our probing calculation tasks. Here, we show layer-wise results of each model in 2-bit and 4-bit integer and floating-point arithmetic expression, seaprately.
}
\label{fig:all_cal}
\vspace{-10pt}
\end{figure*}

Figure \ref{fig:all_cal} shows performances of each size LLaMA 2 model dealing with 1-2bit and 3-4bit integer and floating-point calculation tasks.

\begin{figure*}[]
\centering
\includegraphics[width=1\linewidth]{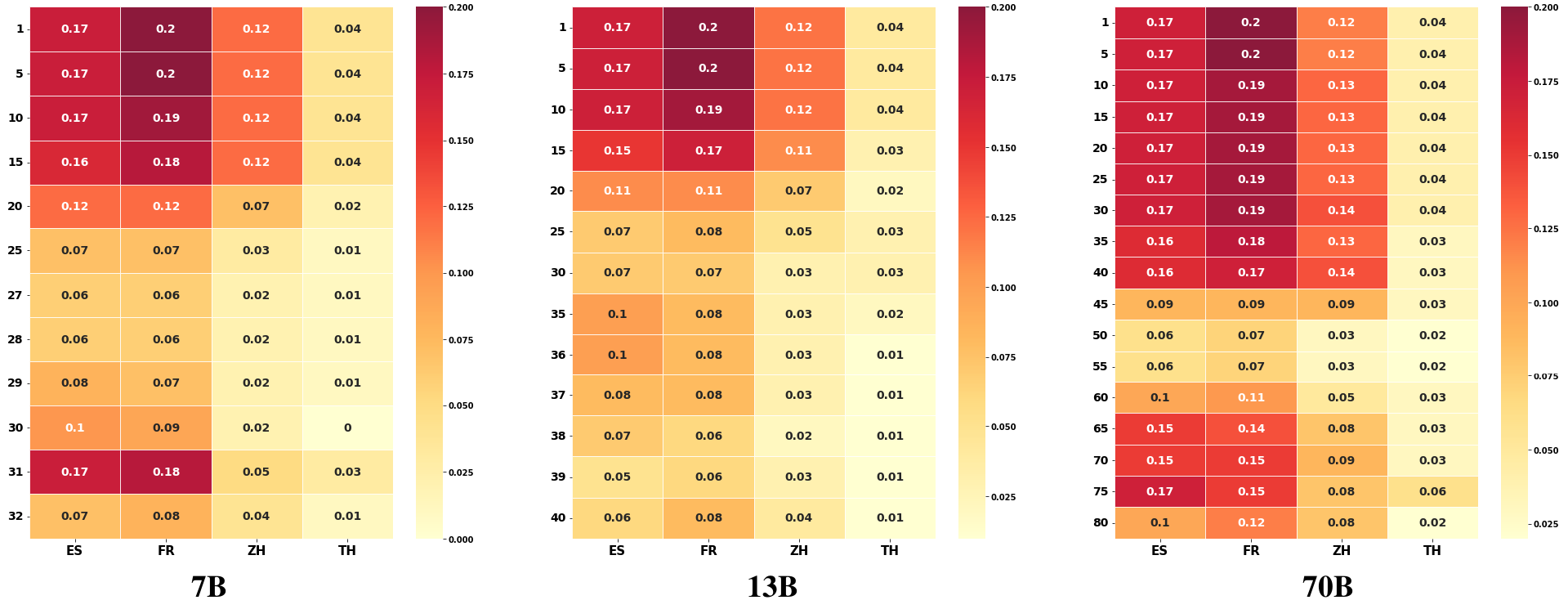}
\caption{Overall Comparison with LLaMA 2-7B to 70B in our probing xMPS-Cal  tasks. 
}
\label{fig:xmps-cal}
\vspace{-10pt}
\end{figure*}

\end{document}